%% file: main.tex
\documentclass[conference]{IEEEtran}
\IEEEoverridecommandlockouts
\usepackage{framed,multirow}

\usepackage{xcolor}
\usepackage{soul}

\usepackage{amssymb}
\usepackage{latexsym}
\usepackage{listings}
\usepackage{graphicx}

\usepackage{hyperref}
\usepackage{xcolor}
\definecolor{newcolor}{rgb}{.8,.349,.1}
\usepackage{bbm}
\newcommand{\indicator}{\mathbbm{1}}

\newcommand{\mytitle}{Domain Adaptation for Holistic Skin Detection}

\definecolor{mygreen}{rgb}{0,0.6,0}
\definecolor{mygray}{rgb}{0.5,0.5,0.5}
\definecolor{mymauve}{rgb}{0.58,0,0.82}

\def\BibTeX{{\rm B\kern-.05em{\sc i\kern-.025em b}\kern-.08em
    T\kern-.1667em\lower.7ex\hbox{E}\kern-.125emX}}
\begin{document}

\title{\mytitle}


\author{
\IEEEauthorblockN{ Aloisio Dourado,
                   Frederico Guth,
                   Teofilo de Campos,
                   Li Weigang
}
\IEEEauthorblockA{
    Universidade de Brasilia (UnB), Departamento de Ci\^encia da Computa\c{c}\~ao - CIC\\
    Campus Darcy Ribeiro, 
    Asa Norte, Brasília - DF \\CEP 70910-900, Brazil\\ 
    \{aloisio.neto, federico.guth\}@aluno.unb.br, t.decampos@oxfordalumni.org, weigang@unb.br}
}

\maketitle

\begin{abstract}
    \input{abstract}
\end{abstract}


\input{body}

\bibliography{defs,refs}
%\bibliography{IEEEabrv,project}{}
\bibliographystyle{IEEEtran}

\end{document}

%% file: abstract.tex
Human skin detection in images is a widely studied topic of Computer Vision for which it is commonly accepted that analysis of pixel color or local patches may suffice. This is because skin regions appear to be relatively uniform and many argue that there is a small chromatic variation among different samples. However, we found that there are strong  biases in the datasets commonly used to train or tune skin detection methods. Furthermore, the lack of contextual information may hinder the performance of local approaches. In this paper we present a comprehensive evaluation of holistic and local Convolutional Neural Network (CNN) approaches on in-domain and cross-domain experiments and compare with state-of-the-art pixel-based approaches. We also propose a combination of inductive transfer learning and unsupervised domain adaptation methods, which are evaluated on different domains under several amounts of labelled data availability. We show a clear superiority of CNN over pixel-based approaches even without labelled
training samples on the target domain. Furthermore, we provide experimental support for the counter-intuitive superiority of holistic over local approaches for human skin detection.

%% file: body.tex
\section{Introduction}

Human skin detection is the task of identifying which pixels of an image correspond to skin. The segmentation of skin regions in images has several applications: video surveillance, people tracking, human computer interaction, face detection and recognition and gesture detection, among many others \cite{shaik_comparative_2015,mahmoodi_sayedi_SkinDetectionSurvey_ijigsp2016}. 

Before the boom of Convolutional Neural Networks (CNNs), most approaches were based on skin-color separation or texture features, as in \cite{huynh-thu_skin-color-based_2002} and  \cite{Shrivastava:2016:CDP:2899522.2899598}. By that time, there were other approaches for image segmentation in general, like  Texton Forest \cite{shotton_semantic_2008} and Random Forest \cite{shotton_real-time_2011}. As occurred with image classification, starting from 2012, convolutional networks have been very successful in segmentation tasks. One of the first approaches using deep learning was patch-based classification \cite{ciresan_deep_2012}, where each pixel is classified using a patch of the original image that surrounds it. This a local approach that does not consider the context of the image as a whole. Later, \cite{shelhamer_fully_2017} introduced the Fully Convolutional Networks (FCNs), a global approach, where image segmentation is done in a holistic manner, achieving state-of-the-art results in several reference datasets. 

%One main difference between local and global approaches is the ground collection process. In local approaches, from one single labelled image, several pixels or patches with corresponding labels can be extracted. On the other hand, in global approaches, as context matters, each labeled image generates one ground truth, thus requiring much more labeled images for training.       

In spite of all the advances that deep fully convolutional neural networks have brought for image segmentation, some common criticism are made to argue that pixel-based approaches are still more suitable for skin detection.  Namely, \label{criticism}
\begin{enumerate}
  \item the need for large training datasets \cite{KAKUMANU20071106}; one may not know in advance the domain of the images that will be used, therefore, no amount of labeled training data may be enough;
  \item the specificity or lack of generalization of neural nets; and 
  \item their prediction time \cite{BRANCATI201733}; especially for video applications where the frame-rate are around 30 or 60 frames-per-second, allowing a maximum prediction time of 17 to 33ms per image. 
\end{enumerate}

Those arguments seem to ignore several proposed approaches that exploit unlabeled data of the domain of interest (unsupervised domain adaptation) or labeled data and models from other domains (inductive transfer learning) to solve the lack of labeled data. Amid the fast evolution of CNNs and domain adaptation techniques, we ask ourselves:  \emph{Do those criticisms still hold for the skin detection problem?}

In this paper, to address the first criticism (on the need of large training datasets), we propose a new Domain Adaptation strategy that combines Transfer Learning and Pseudo-Labeling \cite{Lee-pseudo} in a cross-domain scenario that works under several levels of target domain label availability. We evaluate the proposed strategy under several cross-domain situations on four well-known skin datasets. We also address the other criticisms with a series of comprehensive in-domain and cross-domain experiments. Our experiments show the effectiveness of the proposed strategy and confirm the superiority of FCN approaches over local approaches for skin segmentation. With the proposed strategy we are able to improve the F$_1$ score on skin segmentation using little or no labeled data from the target domain.

Our main contributions are:
\begin{enumerate}
\item the proposal of a new Domain Adaptation strategy that combines Pseudo-Labeling and Transfer Learning for cross-domain training;
\item a comparison of holistic versus local approaches on in-domain and cross-domain experiments applied to skin segmentation with an extensive set of experiments; 
\item a comparison of CNN-based approaches with state-of-the-art pixel-based ones; and
% \item an evaluation of a Domain Adaptation (DA) method to overcome dataset bias with little or no labeled data from target domains; 
\item experimental assessment of the generalization power of different human skin datasets (domains).
\end{enumerate}

\section{Background} \label{sec:background}
% Teo's edition:
%
% Transfer Learning (TL) and Domain Adaptation (DA) are alternatives to
% avoid the cost of acquiring
% large amounts of labeled data to train Deep Convolutional Neural
% Networks in a supervised manner frm scratch. While domain adaptation
% exploits available unlabeled data of the target domain,
% transfer learning uses data and/or models of a similar domain to
% solve a different task.o

% Domain Adaptation is a particular case of transfer learning \cite{csurka_comprehensive_2017}.
In this section we briefly discuss image segmentation using Fully Convolutional Networks (FCN); existing approaches of transfer learning  and 
domain adaptation; and related works on skin segmentation.

\subsection{Fully Convolutional Networks (FCN) for image Segmentation}  % cross-domain pseudo-labels} 
In opposition to patch-based classification \cite{ciresan_deep_2012}, where each pixel is classified using a patch of the original image that surrounds it, the FCN-based approach for image segmentation introduced by \cite{shelhamer_fully_2017} considers the context of the whole image.  The FCNs are Convolutional Neural Networks (CNN) in which all trainable layers are convolutional. Therefore, they can
be quite deep but have a relatively small number of parameters, due to the lack of fully connected layers. Another advantage of FCNs is that, in principle,
the dimensionality of the output is variable and it depends on the dimensionality
of the input data. 

FCNs gave rise to the idea of encoder-decoder architectures, which
have upsampling methods, such as unpooling layers and transpose convolutions (so-called deconvolution layers).
These methods can perform segmentation taking the whole image as an input signal
and generate full image segmentation results in one forward step of the network,
without requiring to break the image into patches. Because of that, FCNs are faster than the patch-based approaches, and overcame the state-of-the-art on PASCAL VOC, NYUDv2, and SIFT Flow datasets, by using Inductive Transfer Learning % \footnote{See discussion on section \ref{sec:background}.}
from ImageNet. 

Following the success of FCNs, \cite{ronneberger_u-net:_2015} proposed the U-Net architecture, that consists of an encoder-decoder structure initially used in biomedical 2D image segmentation. In  U-Net, the encoder path is a typical CNN, where each down-sampling step doubles the number of feature channels.

 What makes this architecture unique is the decoder path, where each up-sampling step concatenates the output of the previous step with the output of the down-sampling with the same image dimensions. This strategy enables precise localization with a simple network that is applied in one shot, rather than using a sliding window.
With this strategy, the U-Net is able to model contextual information, which increases
its robustness and allows it to generate segmentation results with a much finer level of detail.
This strategy is simpler and faster than more sophisticated methods, such as
those that combine CNNs with conditional random fields \cite{arnab_etal_CNN_CRF_pami2018}.
The method of \cite{zheng_etal_CRF_RNN_iccv2015}, which models CRFs as 
recurrent neural networks (CRF-as-RNN), enables a single end-to-end trainining/inference process
for segmentation, generate sharper edges in the segmentation results in comparison to
the standard U-Net. However, CRF-as-RNN is much slower than U-Net due to the nature of RNNs.

The original U-Net architecture does not take advantage of pre-trained classification networks. In order to deal with small amounts of labeled data, the authors made extensive use of Data Augmentation, which has been proven efficient in a many cases \cite{xu_improved_2016,vasconcelos_increasing_2017,perez_effectiveness_2017,wong_understanding_2016}.

Several variations of U-Net have been proposed since then. For example, the  V-Net \cite{milletari_v-net:_2016} is also an encoder-decoder network adapted to segmentation of 3D biomedical images. Nowadays, one of the most used variations consists in replacing the encoder branch with a pre-trained classification network like Inception \cite{szegedy_rethinking_2016} or Resnet \cite{he_etal_ResNet_cvpr2016}, combining the U-Net architecture with the original approach of Fully Convolutional Networks. Another common strategy is the use of short-range residual connections in the convolutions blocks of the encoder and decoder branches of U-Net, as in \cite{DBLP:journals/corr/abs-1805-09233}. Due to its simplicity and high performance, we choose U-Net as the FCN reference model in our experiments. 

As Deep Neural Networks require high amounts of labeled data to be trained,
Transfer Learning and Semi-Supervised Learning
learning methods can be employed to dramatically reduce
the cost of acquiring training data.  
While semi-supervised learning exploits available unlabeled
data in the same domain, transfer learning is a family of methods
that deal with change of task or change of domain.
Domain Adaption (DA) is a particular case of transfer
learning \cite{csurka_comprehensive_2017}.
% In this section we will
% give some background on existing approaches of semi-supervised
% learning and domain adaptation, focusing on Visual applications,
% eespecially image segmentation.
We will discuss these methods in the next sub sections.

\subsection{Transfer Learning Base Concepts}  % cross-domain Transfer 

Here we present Transfer Learning (TA) and Domain Adaptation (DA) concepts that will be used in the rest of the paper. We follow the notation of \cite{pan_survey_2010}.

A domain $\mathcal{D}$ is composed of a $d$-dimensional feature space $\mathcal{X}\subset R^d$ with a marginal probability distribution $P(\mathrm{X})$. A task $\mathcal{T}$ is defined by a label space $\mathcal{Y}$ with conditional probability distribution $P(\mathrm{Y}|\mathrm{X})$.

In a conventional supervised machine learning problem, given a sample set $\mathrm{X}=\{\mathrm{x_1}, \cdots, \mathrm{x_n}\} \in \mathcal{X}$ and the corresponding labels $\mathrm{Y}=\{\mathrm{y_1}, \cdots, \mathrm{y_n}\} \in \mathcal{Y}$, $P(\mathrm{Y}|\mathrm{X})$ can be learned from feature-label pairs in the domain. % \cite{csurka_comprehensive_2017}.
%% parágrafo abaixo pode ser removido se precisarmos de espaço
Suppose we have a source domain $\mathcal{D}^s=\{\mathcal{X}^s, P(X^s)\}$ with a task $\mathcal{T}^s=\{\mathcal{Y}^s, P(Y^s|X^s)\}$ and a target domain $\mathcal{D}^t=\{\mathcal{X}^t, P(X^t)\}$ with a task $\mathcal{T}^t=\{\mathcal{Y}^t, P(\mathrm{Y}^t|\mathrm{X}^t)\}$. If the two domains correspond ($\mathcal{D}^s=\mathcal{D}^t$) and the two tasks are the same ($\mathcal{T}^s=\mathcal{T}^t$), we can use conventional supervised Machine Learning techniques. %, using $\mathcal{D}^s$ as a training set and $\mathcal{D}^t$ as a test set \cite{csurka_comprehensive_2017}.
Otherwise, adaptation and/or transfer methods are required.

If the source and target domains are represented in the same feature space ($\mathcal{X}^s = \mathcal{X}^t$), but with different probabilitiy distributions ($P(\mathrm{X^s}) \neq P(\mathrm{X}^t)$) due to domain shift or selection bias, the transfer learning problem is called homogeneous.
If ($\mathcal{X}^s \neq \mathcal{X}^t$), the problem is heterogeneous TL~\cite{csurka_comprehensive_2017,pan_survey_2010}.
In this paper, we deal with homogeneous transfer learning as we use the same feature space representation for source and target datasets.
% If the two domains have different data representations, it is called heterogeneous TL \cite{csurka_comprehensive_2017}. 

Domain Adaptation is the problem where tasks are the same, but data representations are different
or their marginal distributions are different (homogeneous). 
% due to distribution mismatch or selection bias (homogeneous transductive TL).
%% Nao eh necessario falar de unsupervised TL se nos nao vamos usar. Isso eh frequentemente confundido com unsupervised DA:
% Unsupervised TL is the case where both,  tasks and the domains, are different but related \cite{pan_survey_2010}.  
% 
% DA methods are unsupervised (also known as transductive TL)
Mathematically, $\mathcal{T}^s = \mathcal{T}^t$ and $\mathcal{Y}^s = \mathcal{Y}^t$,
but $P(\mathrm{X}^s) \neq P(\mathrm{X}^t)$.
%Therefore the tasks are the same, the set of labels are the same, but $P(\mathrm{Y}|\mathrm{X}^t)$ may be different from $P(\mathrm{Y}|\mathrm{X}^s)$.

Most of the literature on domain adaptation for visual applications is dedicated to image classification \cite{csurka_comprehensive_2017}.
To extend the domain adaptation concepts to the image segmentation problem, we treat the Skin Segmentation as a pixel-wise classification problem.
%, where to each pixel of the image is assigned a label.

\subsection{Inductive Transfer Learning}

When source and target domains are different ($\mathcal{D}^s\neq \mathcal{D}^t$), models trained on  $\mathcal{D}^s$ may not perform well while predicting on $\mathcal{D}^t$ and if tasks are different ($\mathcal{T}^s\neq \mathcal{T}^t$), models trained on $\mathcal{D}^s$ may not be directly applicable on $\mathcal{D}^t$. Nevertheless, when $\mathcal{D}^s$ maintains some kind of relation to $\mathcal{D}^t$ it is possible to use some information from  $\{\mathcal{D}^s,\mathcal{T}^s\}$ to train a model and learn $P(\mathrm{Y}^t|\mathrm{X}^t)$ through a processes that is called Transfer Learning (TL) \cite{pan_survey_2010}.

According \cite{csurka_comprehensive_2017}, the Transfer Learning approach is called inductive if the target task is not exactly the same as the source task, but the tasks are in some aspects related to each other. For instance, consider an image classification task on ImageNet~\cite{russakovsky_etal_ILSVRC_ijcv2015} as source task and a Cats vs Dogs classification problem as a target task. If a model is trained on a dataset that is as broad as ImageNet, one can assume that
most classification tasks performed on photographies downloaded from the web are 
subdomains of ImageNet which includes the Cats vs Dogs problem (i.e.\ $\mathcal{D}^{\tt cats\times dogs} \subset    \mathcal{D}^{\tt ImageNet}$), even though the tasks are different ($\mathcal{Y}^{\tt ImageNet} = \mathbb{R}^{1000}$ and
$\mathcal{Y}^{\tt cats\times dogs} = \mathbb{R}^2$). 
This is the case of a technique to speed up convergence in Deep CNNs that became popularised as {\em Fine Tuning} for vision applications.

In deep artificial neural networks, fine tuning is done by taking a pre-trained model,
modifying its final layer so that its output dimensionality matches $\mathcal{Y}^t$ and
further training this model with labelled samples in $\mathcal{D}^t$.

Further to fine tuning, a wide range of techniques has been proposed for inductive TL~\cite{pan_survey_2010}, particularly using shallow methods, such as SVMs~\cite{Aytar_ICCV2011}, where the source domain is used to regularize the learning process. The traditional fine tuning processes usually requires a relatively large amount of labeled data from the target domain \cite{csurka_comprehensive_2017}. In spite of that, this technique is very popular with CNNs. In this work, we compare our proposed domain adaptation approach to inductive transfer learning applied to the skin segmentation problem. Details of the inductive transfer learning setup used in our experiments are presented on Section~\ref{subsec:tl}.

\subsection{Unsupervised Domain Adaptation}

Domain adaptation methods are called unsupervised (also known as transductive TL) when labeled data is available only
on source domain samples.
% We will also focus on homogeneous Domain Adaptation, where both domains share the same feature space.
%
% Unsupervised TL is the case where both,  tasks and the domains, are different but related \cite{pan_survey_2010}.  

Several approaches have been proposed for unsupervised DA, most of them
were designed for shallow learning methods \cite{csurka_DA_book2017}.
The methods that exploit labeled samples from the source domain follow
a similar assumption to that of Semi-Supervised Learning methods,
with the difference that test samples come from a new domain.
This is the case of \cite{JDA_ICCV2013} and
\cite{farajidavar_etal_ATTM_bookchapter2017}.
Both methods start with a standard supervised
learning method trained on the source domain in order to classify samples
from the target domain. The classification results are taken as pseudo-(soft)labels
and used to iteratively improve the learning method in a way that it
works better on the target domain. 

When labeled samples are not available at all, it is possible to perform
unsupervised transfer learning using methods that perform feature space transformation.
Their goal is to align source and target domain samples to minimise the
discrepancy between their probability density functions \cite{Gretton_2006}.
Style transfer techniques such as that of \cite{Gatys_etal_StyleTransfer_cvpr2016}
achieve a similar effect, but their training process is much more complex.

\subsection{Semi-supervised learning}  % cross-domain pseudo-labels} 
\label{subsec:SSL}

Semi-supervised learning methods deal with the problem in which
not all training samples have labels \cite{zhu_semiSupervisedSurvey2005,murphy_book2012}.
Most of these methods use a density model in order to propagate labels
from the labeled samples to unlabeled training samples. This step is usually
combined with a standard supervised learning step in order
to strengthen the classifiers, c.f.\ \cite{Leistner_Saffari_Santner_Bischof_2009,criminisi_shotton_SemiSupervisedForests_2013}.

There are several semi-supervised learning approaches for deep neural networks. Methods include training networks using a combined loss of an auto-encoder and a classifier \cite{ranzato_semi-supervised_2008}, discriminative restricted Boltzmann machines \cite{larochelle_classification_2008} and  semi-supervised embeddings \cite{Weston:2008:DLV:1390156.1390303}.

\cite{Lee-pseudo} proposed a simple yet effective approach, known as Pseudo-Labelling, where the network is trained in a semi-supervised way, with labeled and unlabeled data in conjunction. During the training phase, for the unlabeled data, the class with the highest probability (pseudo-label) is taken as it was a true label. To account for the unbalance between true and pseudo labels, the loss function uses a balancing coefficient to adjust the weight of the unlabeled data on each mini-batch. As a result, pseudo-label works as an entropy regularization strategy.

These methods assume that training and test samples belong to the same domain, or at least that they are very similar 
($\mathcal{D}^s\approx\mathcal{D}^t$). 
The original Pseudo-Labelling method only considers in-domain setups. As we are dealing with homogeneous DA, in this work we extend Pseudo-Labeling to cross-domain, as shown in  Section~\ref{subsec:pseudo}. We evaluate this extension under several target label availability constrains, from semi-supervised learning (when some target labels are available) to fully unsupervised (when no target labels are available).

\subsection{Related Works on Skin Detection}

\cite{BRANCATI201733} achieved state-of-the-art results in skin segmentation using correlation rules between the YCb and YCr subspaces to identify skin pixels on images. A variation of that method was proposed by \cite{Faria2018}, who claimed to have achieved a new state-of-the-art plateau on rule-based skin segmentation based on neighborhood operations.  \cite{Lumini2018FairCO} compared different color-based and CNN based skin detection approaches on several public datasets and proposed an ensemble method.

In contrast to Domain Adaptation for image classification, it is difficult to find literature focused on domain adaptation methods for image segmentation \cite{csurka_comprehensive_2017}, especially for the skin detection problem.  \cite{SANMIGUEL20132102} use agreement of two detectors based on skin color thresholding, applied to selected images from several manually labeled public datasets for human activity recognition, but do not explore their use in cross-domain setups. \cite{Conaire4270446} also use two independent detectors, with their parameters selected by maximising agreement on correct detections and false positives to  dynamically change a classifier on new data automatically without any user annotation.  \cite{Kamnitsas/10.1007/978-3-319-59050-9_47} use unsupervised domain adaptation to improve brain lesion detection in MR images.  \cite{Bousmalis2017UnsupervisedPD} developed a generative adversarial network based model which adapts source-domain images to appear as if drawn
from the target domain, a technique that enables dataset augmentation for several computer vision tasks.

In this work we compare two CNN approaches (one patch-based and one fully convolutional) with above mentioned state-of-the-art pixel-based methods for in-domain skin detection. We also compare the two CNN approaches to each other in cross-domain setups, even in the absence of target-domain labeled data. Unfortunately, previous state-of-the-art pixel-based skin segmentation papers do not present results on cross-domain setups. 

% \hl{As expected, CNN solutions surpassed color based methods, but with the drawback of requiring a substantial amount of labeled data for training. To face this drawback, we propose and evaluate a new DA approach that also overcame color-based methods, however,  without using any labeled data from target dataset.}

\section{Methods}

In this paper, we compare CNN patch-based and holistic approaches to the state-of-the-art pixel-based approaches on skin segmentation using in-domain training. We also propose to combine the strengths of both inductive transfer learning and unsupervised or semi-supervised domain adaptation using Pseudo-Labeling in order to address the lack of training data issue using cross-domain setups.  In this section we present details of the training approaches, models and experimental protocols.

% The details of evaluated approaches, models and experiment protocols are presented in this section.

\subsection{Cross-domain training approaches} 
In order to exploit domain adaptation techniques to address training data avaiability problem for skin segmentation, we evaluate conventional transductive transfer learning using fine tuning, our cross-domain extension applied to the Pseudo-Labeling approach of \cite{Lee-pseudo} and our proposed combined approach that uses both inductive transfer learning and unsupervised or semi-supervised DA. Here we present each one of these training approaches.

\subsubsection{Inductive Transfer Learning approach} 
\label{subsec:tl}
\begin{figure}
  \centering
    \includegraphics[width=.48\textwidth, trim={8cm 6cm 8cm 6cm},clip]{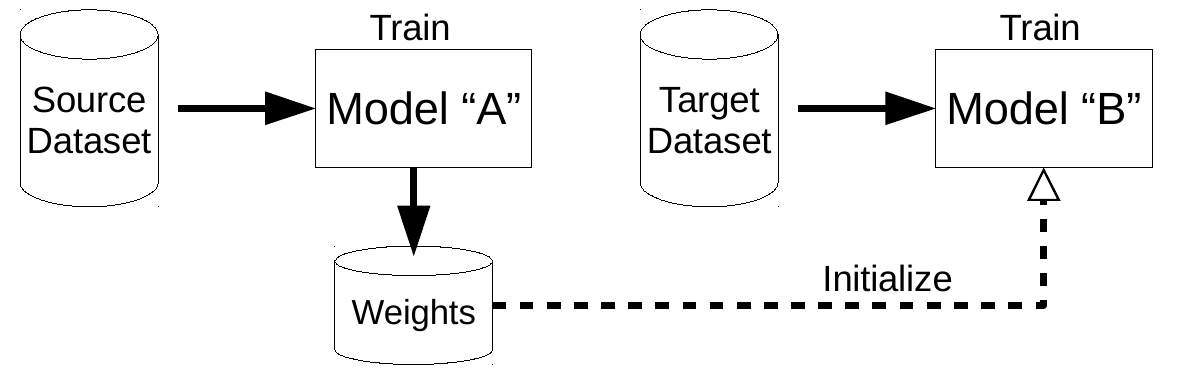}
  \caption{Inductive Transfer Learning by ``fine tuning'' parameters of a model to a new domain. Model ``A'' parameters are trained on the source dataset. Model ``B'' parameters are initialized from Model ``A'' parameters. Model ``B'' is then ``fine tuned'' to the new domain.}
  \label{fig:fine-tune}
\end{figure}

For inductive transfer learning with deep networks, we use the learnt parameters from the
source domain as starting point for optimisation of the parameters of the
network on the target domain. % (``fine tuning'').
The optimisation first focuses on the modified output layer, which is
intimately linked with the classification task. Other layers are initially
frozen, working as a feature extraction method. Next, all parameters are
unfrozen and optimisation carries on until convergence. This can be seen
as a way to regularise the learnt parameters on the source domain.
Figure~\ref{fig:fine-tune} illustrates this process.

% \subsection{Unsupervised Domain Adaptation}
\subsubsection{Cross-domain Pseudo Labeling approach} 
\label{subsec:pseudo}
In this work, we propose a method that relates the pseudo-label approach of \cite{Lee-pseudo}, but instead of using the same model and domain for final prediction and pseudo-label generation, we use a model trained in a different domain to generate pseudo labels for the target domain. These pseudo-labels are then used to fine-tune the original model or to train another model from scratch in a semi-supervised manner. We call this technique \textbf{cross-domain pseudo-labelling}.

% \hl{Now this is redundant:}
% Our proposed ``cross-domain pseudo-label'' approach is a variation of \cite{Lee-pseudo} that uses predictions of a model trained in another domain as pseudo-labels. These pseudo-labels are used, in conjunction with labeled data of the target domain, to train a final model.

\begin{figure}
  \centering
    \includegraphics[width=.48\textwidth, trim={8cm 5.75cm 8cm 5.75cm},clip]{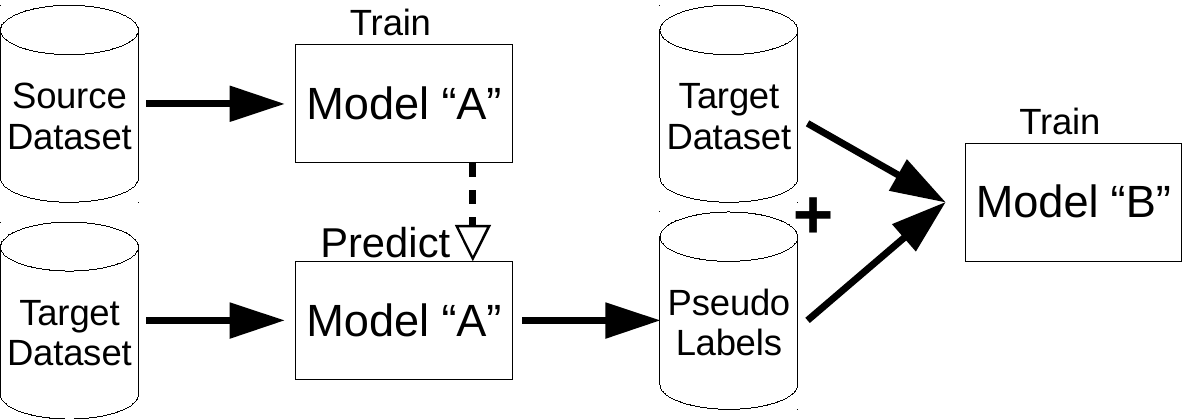}
  \caption{Semi-supervised and unsupervised Domain Adaptation by cross-domain pseudo-labelling. Model ``A'' is trained on the source dataset and it is used to predict labels on the target dataset. Then, the target dataset and previously predicted labels are used to train Model ``B''. When no labels are available on the target dataset, the process is fully unsupervised.}
  \label{fig:pseudo}
\end{figure}

Figure \ref{fig:pseudo} illustrates this procedure. This approach allows us to train the final model with very few labeled data of the target domain. In the worst case scenario, the model can be trained with no true label at all, in a fully unsupervised fashion. This still takes advantage of Entropy Regularization of the pseudo-label technique. 

\subsubsection{Combined approach}
\label{subsec:combined}
\begin{figure}
  \centering
    \includegraphics[width=.48\textwidth, trim={8cm 3.5cm 8cm 3.5cm},clip]{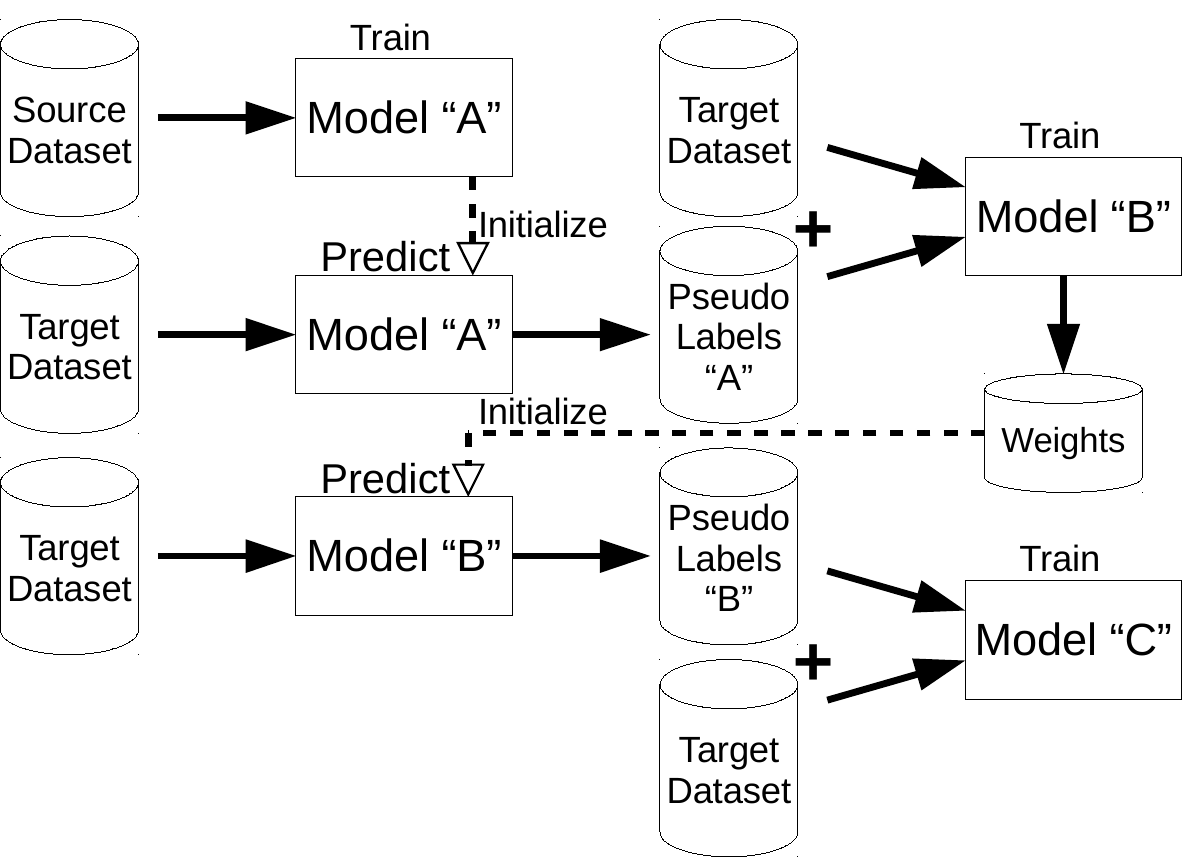}
    \caption{Combined transfer learning and domain adaptation approach. Model ``A'' is trained on the source dataset and it is used to predict labels on the target dataset. Then, target dataset and previously predicted labels are used to train Model ``B'' which is fine tuned on the target dataset before be used to generate a new set of more accurate pseudo-labels.}%  Because fine tuning requires at least a small amount of real labels, this processes is semi-supervised.}
  \label{fig:combined}
\end{figure}

Our last approach consists in combining fine tune and pseudo labeling approaches in order to improve final model performance. Figure \ref{fig:combined} illustrates this procedure.
We use weights obtained from a cross-domain pseudo-label model (Model ``B'') to fine tune a model that will be used to generate a more accurate set of pseudo-labels.
These new pseudo-labels are then used in one in-domain pseudo-label training round to get the final model (``Model C'').
 The intuition behind this approach is that using a more accurate set of labels jointly with weights of a better model should lead to better results. Because of the fine tuning step, which requires at east some labels from the target dataset, this approach is semi-supervised. 

\subsection{Models}
We evaluated two approaches for skin segmentation, a local (patch-based)
convolutional classification method and a holistic (FCN) segmentation method. Here we describe these methods.

% Excellent segmentation tutorial: https://www.jeremyjordan.me/semantic-segmentation/

\subsubsection{Patch-based CNN}
The patch-based approach uses the raw values of a small region of the
image to classify each pixel position based on its neighbourhood.
Inspired by the architecture described by \cite{ciresan_deep_2012},
we use a 3 convolutional layer network with max pooling between
convolutions, but we add ReLU activation function in the inner layers.
As input, we use a patch of $35\times35$ pixels and 3 channels,
to allow the network to capture the surroundings of the pixel.
This patch size is similar to that used by \cite{ciresan_deep_2012}
($32\times32$), but we chose an odd number to focus the prediction
in the center of the patch.
The output of the network consists of two fully connected layers
and a sigmoid final activation for binary classification.
% \hl{[Insert diagram of this architecture?]}
For this approach, the images are not resized.
To reduce the cost of training while maintaining data diversity, data subsampling
is used so that only $512$ patches are randomly selected from each image.
For prediction, all patches are extracted in a sliding window fashion,
making one prediction per pixel.
Due to the path size, the prediction processes generates a 17 pixels wide border
where this method does not predict an output, so zero padding is applied.
This does not harm the predictions, since the presence
of skin near the borders is rare in all datasets used.

\subsubsection{Holistic segmentation FCN}

\begin{figure}
  \centering
    \includegraphics[width=.49\textwidth, trim={0cm 0cm 0cm 0cm},clip]{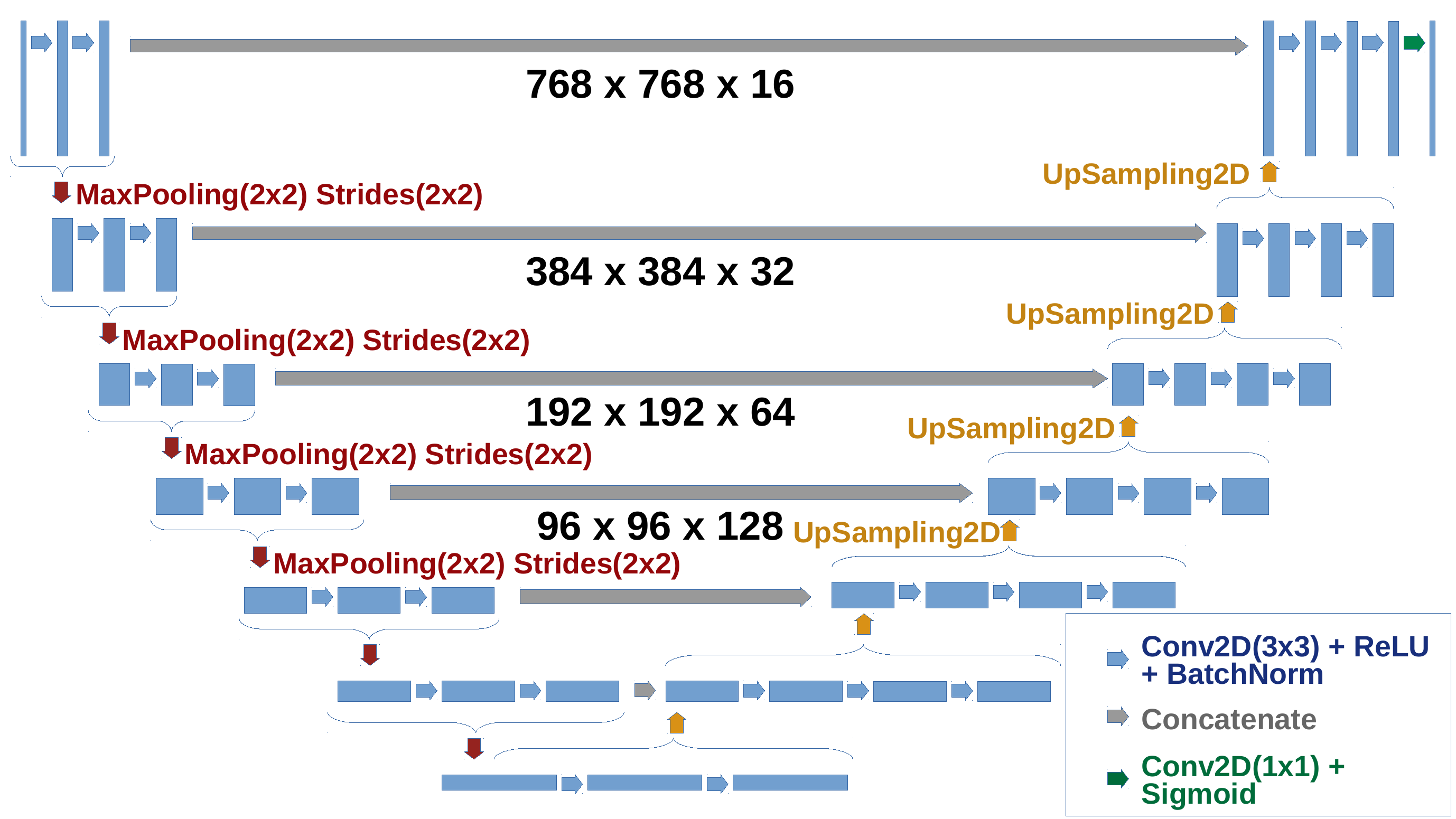}
  \caption{Our variation of the U-Net architecture for holistic image segmentation.}
  \label{fig:unet}
\end{figure}

Due to its simplicity and performance, we choose to use the U-Net as the holistic segmentation method to be evaluated
in this paper.
Our model follows the general design proposed by \cite{ronneberger_u-net:_2015},
but we use a 7-level structure with addition of batch normalization
between the convolutional layers, as shown in figure~\ref{fig:unet}.
We also use an input frame of $768\times768$ pixels and 3 channels to fit most images,
and same size output.  

Smaller images are framed in the center of the input and larger ones are resized in
a way that its larger dimension fits the input frame. For evaluation purposes,
predictions are done over the images restored to their original sizes.

\subsection{Evaluation measures and loss function}
\label{sec:loss}
From the literature, we have identified that the the most popular
evaluation criteria for image segmentation are: Accuracy (Acc), Jaccard Index (a.k.a.\ Intersection Over Union, IoU), Precision, Recall and F$_1$ Score (a.k.a.\ S{\o{}}rensen–Dice Coefficient or Dice Similarity Coefficient, DSC). In this section, we revise them following a notation that helps to compare them.
For each given class label, let $\vec{p} \in [0,1]^\mathcal{I}$ be the vector of predicted probabilities for each pixel (where $\mathcal{I}$ is the number pixels in each image), $\vec{q} \in \{0,1\}^\mathcal{I}$ be the binary vector that indicates, for each pixel, if that class has been detected, based on $\vec{p}$, and $\vec{g}$ be the ground truth binary vector that indicates the presence of that label on each pixel.
We have the following definitions:
\begin{eqnarray} \label{eq:acc}
  \textrm{Acc} & = & \frac{\sum_i^\mathcal{I} \indicator_{g_i}(q_i)}{\mathcal{I}} = \frac{\vec{q}\cdot\vec{g} + (\vec{1}-\vec{q})\cdot(\vec{1}-\vec{g})}{\mathcal{I}}
  \\
 \label{eq:iou}
  \textrm{IoU} & = & \frac{|\vec{q} \cap \vec{g}|}{|\vec{q} \cup \vec{g}|} = \frac{\vec{q} \cdot \vec{g}}{\sum_i^\mathcal{I}{\max{(p_c, g_c)}}} = \frac{\vec{q} \cdot \vec{g}}{|\vec{q}| + |\vec{g}| - \vec{q}\cdot\vec{g}}
  \\
  \textrm{Prec} & = & \frac{\vec{q}\cdot\vec{g}}{|\vec{q}|}
  \\
  \textrm{Rec} & = & \frac{\vec{q}\cdot\vec{g}}{|\vec{g}|}
  \\
  \label{eq:f1}
  \textrm{F}_1 & = & \left(\frac{\textrm{Prec}^{-1} + \textrm{Rec}^{-1}}{2}\right)^{-1} = 2 \cdot \frac{\vec{q} \cdot \vec{g}}{|\vec{p}| + |\vec{g}|}
\end{eqnarray}
Also, from \ref{eq:iou} and \ref{eq:f1}, we can derive that the Jaccard index and F$_1$ score are monotonic in one another:
\begin{eqnarray} 
  \textrm{IoU} & = & \frac{\textrm{F}_1} {2-\textrm{F}_1} \hspace{3em}  \therefore  \textrm{F}_1  =  \frac{2 \cdot \textrm{IoU}}{1 + \textrm{IoU}} 
\end{eqnarray}
As such, there is no quantitative argument to prefer one over the other. Qualitatively, though, we recommend using F$_1$ score as it is a more prevalent metric in other fields.  Although accuracy has been widely used, we consider that not to be a good metric, as its numerator not only considers true positives, but also true negatives, and a null hypothesis gives high accuracy on imbalanced datasets.

As for the loss function, training objective and evaluation metric should be as close as possible, but F$_1$ score is not differentiable.  Therefore, we used a modified (and differentiable) S{\o{}}rensen–Dice coefficient,
given by equation \ref{eq:a}, where $s$ is the smoothness parameter that was set to $s=10^{-5}$. The derived loss function is given by (\ref{eq:b}).

\begin{equation} \label{eq:a}
 softDiceCoef(\vec{p},\vec{g}) = \frac{s + 2 \vec{p} \cdot \vec{g}}{s + |\vec{p}| + |\vec{g}|}
\end{equation}
\begin{equation} \label{eq:b}
 DiceLoss(P,G) = 1 - softDiceCoef(P,G)
\end{equation}

\subsection{Data augmentation}
In both local and holistic models, the image pixels are normalized to 0 to 1 and the sigmoid activation function applied to the output. In both models we also used data augmentation, randomly varying pixels values in the HSV colour space. For the U-Net model we also used random shift and flip.

\section{Experiments and results}

The main goal of our experiments is to evaluate the performance of homogeneous transductive fine-tuning, cross-domain pseudo-labelling, and a combined approach in several domains and under different availability of labeled data on the target domain. 
To achieve this goal, we used four well-known datasets dedicated to skin segmentation (described in Section~\ref{sec:datasets}) and permuted them as source and target domain. The first set of experiments (Section~\ref{sec:indomain}) was conducted to compare the CNN approaches to the state-of-the-art pixel-based works. The second set of experiments (Section~~\ref{sec:baseline}) was designed to evaluate the generalization power and the amount of bias in each dataset. Next, in order to evaluate the cross-omain approaches, for each pair of datasets and for each approach we performed a range experiments using different amounts of labeled training data from the target domain (Section~\ref{sec:da_results}).

\subsection{Datasets}
\label{sec:datasets}
The datasets we used were Compaq \cite{compaq} -- a very traditional skin dataset with 4,670 images of several levels of quality;
SFA \cite{casati_sfa} --  a set of 1,118 face images  obtained from two distinct datasets, some of them with white background; Pratheepan \cite{pratheepan} -- 78 family and face photos, randomly  downloaded using Google; and VPU \cite{SANMIGUEL20132102} -- 290 images extracted from video surveillance cameras.

In order to evaluate the methods, \cite{SANMIGUEL20132102} proposed a pixel-based split of trainining and testing samples (not image based) for the VPU dataset, making it impossible to evaluate holistic methods. The other datasets do not have a standard split of samples. For this reason, we adopted the same test split reported by the authors of SFA \cite{casati_sfa}, which uses 15\% of the images for testing and the remaining for training on all these datasets.

As discussed in Section \ref{sec:loss}, most works on Skin Segmentation report their results in terms of Precision (Prec), Recall and F$_1$ score.
So, we use these metrics while comparing ours results with others. In these situations we also provide Accuracy (Acc) and Intersection over Union (IoU).
When comparing results of our own approaches, in more dense tables, we just present results in terms of F$_1$ score.

\subsection{In-domain evaluations}
\label{sec:indomain}

The same-domain training evaluation results are shown on tables \ref{tab:1}, \ref{tab:2}, \ref{tab:3} and \ref{tab:4}. Our fully convolutional U-Net model surpassed all recent works on skin segmentation available for the datasets in study, and, in most of the cases, our patch-based CNN model stands in second, confirming the superiority of the deep learning approaches over color-based ones.
The results also show that the datasets have different levels of difficulty, being VPU the most challenging one and SFA the least challenging one. 
The best accuracy was obtained on VPU, but this is because this is a heavily unbalanced dataset where most pixels belong to background.
As for all remaining criteria, the best results occured on SFA, which confirms our expectation, as SFA is a dataset of frontal mugshot style photos.

%SFA-SFA RESULTS
\begin{table}[t]
  \caption{Same domain results on the SFA dataset (in \%).} 
  \label{tab:1}
  \centering
  \begin{tabular}{|c|c|c|c|c|c|}
    \hline
    Model & Acc & IoU & Prec & Recall & F$_1$\\
    \hline
    % \cite{Faria2018}
    Faria (2018) &-& -& 92.88 & 39.58 & 55.51 \\
    Our p-based  &91.14& 82.17& 89.71& 91.00& 90.35 \\
    \textbf{Our U-Net} &\textbf{97.94}& \textbf{92.80}& \textbf{96.65}& \textbf{95.89}&\textbf{96.27}  \\
    \hline
  \end{tabular}
\end{table}

%Compaq-Compaq results
\begin{table}[t]
  \caption{Same domain results on the Compaq dataset (in \%).} 
  \label{tab:2}
  \centering
  \begin{tabular}{|c|c|c|c|c|c|}
    \hline
    Model & Acc & IoU & Prec & Recall & F$_1$\\
    \hline
    % \cite{BRANCATI201733}
    Branc.(2017)&-& -& 43.54 & \textbf{80.46} & 56.50 \\
    Our p-based  &90.18& 46.00& 58.92& 73.59&65.45 \\
    \textbf{Our U-Net}  &\textbf{92.62}& \textbf{54.47}& \textbf{68.49}& 71.64& \textbf{70.03}\\
    \hline
  \end{tabular}
\end{table}

%Pratheepan-Pratheepan results
\begin{table}[t]
  \caption{Same domain results on the Pratheepan dataset (in \%).} 
  \label{tab:3}  
  \centering
  \begin{tabular}{|c|c|c|c|c|c|}
    \hline
    Model & Acc & IoU & Prec & Recall & F$_1$\\
    \hline
    Branc.(2017) % \cite{BRANCATI201733}
    &-& -& 55.13 & 81.99 & 65.92 \\
    Faria (2018)% \cite{Faria2018}
    &-& -& 66.81 & 66.83 & 66.82 \\
    Our p-based  &87.12& 55.57& 59.83& \textbf{82.49}&69.36 \\
    \textbf{Our U-Net}  &\textbf{91.75}& \textbf{60.43}& \textbf{72.91}& 74.51& \textbf{73.70}\\
    \hline
  \end{tabular}
\end{table}

%VPU-VPU results
\begin{table}[t]
\caption{Same domain results on the VPU dataset (in \%).} 
\centering
\begin{tabular}{|c|c|c|c|c|c|}
\hline
Model & Acc & IoU & Prec & Recall & F$_1$\\
\hline
SMig.(2013) % \cite{SANMIGUEL20132102}
&-& -& 45.60 & \textbf{73.90} & 56.40 \\
Our p-based  &93.48& 14.14& 46.34& 42.82& 44.51\\
\textbf{Our U-Net}  &\textbf{99.04}& \textbf{45.29}& \textbf{57.86}& 71.33& \textbf{63.90}\\
\hline
\end{tabular}
\label{tab:4}
\end{table}

\subsection{Cross-domain baseline results}
\label{sec:baseline}

The cross-domain capabilities of our models and generalization power of domains are shown on table \ref{tab:5}, which presents source only mean F$_1$ scores results without any transfer or adaptation to target dataset. As we can see, source dataset Compaq in conjunction with the U-Net Model presented the best generalization power on targets SFA and Pratheepan. Source dataset Pratheepan also in conjunction with the U-Net Model did better on targets Compaq and VPU.  These source-only setups surpassed the respective color-based approaches shown on previous tables, except for the VPU dataset.

Note that the patch-based model surpassed U-Net when using source domains with low generalization power like SFA and VPU. For example, using VPU as source domain and SFA as target, patch-based reached mean F$_1$ score of 82.63\%, while U-Net only got 14.83\%. Using SFA as source and Compaq as target, patch-based also surpassed U-Net (54.80\% vs.\ 18.92\%). 
These results are expected, since SFA and VPU are datasets of very specific domains with little variation in the type of scenes between their images (SFA images are close-ups on faces and VPU images are typical viwes from conference rooms or surveillance cameras). On the other hand, Compaq and Pratheepan include images with a wide range of layouts. Therefore, SFA and VPU only offer relevant information at a patch level for skin detection, their contexts are very specific, which hinders their generalisation ability. If the goal is to design a robust skin detector and avoid negative transfer, our results show that it is better to use Compaq or Prateepan as source samples. 

%Cross-domain results
\begin{table}[tb]
\caption{Cross-domain mean F$_1$ scores (\%) obtained without transfer nor adaptation.} 
\centering
\begin{tabular}{|c|c|c|c|c|c|}
\hline
\multirow{2}{*}{Model} & Source & \multicolumn{4}{c}{Target Domain} \vline\\
\cline{3-6}
 & Domain & SFA & Compaq & Prathee. & VPU\\
\hline
\multirow{4}{*}{U-net}&SFA& -& 18.92 & 44.98 & 11.52 \\
 &Compaq& \textbf{86.14}& -& \textbf{75.30}& 23.67\\
  &Prathee.& 80.66& \textbf{63.49}& -& \textbf{36.68}\\
  &VPU& 14.83& 44.71& 48.02& -\\
\hline
\multirow{4}{*}{Patch}&SFA& -& 54.80 & 62.92 & 21.60 \\
 &Compaq& 71.28& -& 72.59& 19.94\\
  &Prathee.& 80.04& 62.68& -& 13.74\\
  &VPU& 82.63& 51.48& 58.34& -\\
\hline
\end{tabular}
\label{tab:5}
\end{table}

\subsection{Domain Adaptation Results}
\label{sec:da_results}
Following the recommendation in the previous section, 
we performed domain adaptation experiments using Compaq and Pratheepan as source datasets. 
Table~\ref{tab:6} presents the F$_1$ scores obtained by the methods and settings we evaluated. For each source$\rightarrow$target pair, we indicate in bold face which result was better than the target-only method. We evaluated the effect of the amount labeled target samples given and present results ranging from no labels (0\%), i.e. an unsupervised domain adaptation setting, to all labels (100\%) given in the target training set, i.e., an inductive transfer set up.  Target only results are provided for comparison purposes, i.e, within domain experiments with the number of training labels ranging from 5 to 100\%. The target only results are expected to be an upper bound in performance when 100\% of the training labels are used because there is no domain change, but they may suffer from the reduced training set size in comparison to the domain adaptation settings.

%##############################################
%Final Table
%##############################################
\begin{table*}[tb]
\caption{U-Net mean F$_1$ scores under different scenarios and domain adptation approaches.} 
\centering
\begin{tabular}{|c|c|c|c|c|c|c|c|}
\hline
\multirow{2}{*}{Source} & \multirow{2}{*}{Target} & \multirow{2}{*}{Approach} & \multicolumn{5}{c}{ Target Training Label Usage} \vline\\
\cline{4-8}
{} & {} & {} & 0\% & 5\% & 10\% & 50\%& 100\%\\
\hline
&SFA& \multirow{4}{*}{Target only} & - & 93.49 & 94.50 & 95.72& \textbf{96.27}\\
Target & Compaq&     {} & - & \textbf{66.84} & \textbf{67.78} & \textbf{69.37}& 70.03\\
only &Pratheepan& {} & - & 46.36 & 59.86 & 69.04& 73.70\\
{}&VPU&        {} & - & 41.27 & 53.44 & 63.18& 63.90\\
\hline
\multirow{12}{*}{Compaq}&\multirow{4}{*}{SFA}& Source only & 86.14 & - & - & -& -\\
{}&{}& Fine-tuning only & - & 92.89 & 94.04 & \bf{95.86}& 95.98\\
{}&{}& Cross-domain pseudo-label only & 88.80 & 88.90 & 89.69&  93.22& -\\
{}&{}& Combined approach & \textbf{89.24} & 90.05 & 90.36 & 94.57& -\\
\cline{2-8}
{}&\multirow{4}{*}{Pratheepan}& Source only & 75.30 & - & - & -& -\\
{}&{}& Fine-tuning only & - & 72.52 & 74.69 & 76.47& \textbf{77.16}\\
{}&{}& Cross-domain pseudo-label only & 75.58 & 75.52 & 77.18 & \textbf{80.08} & -\\
{}&{}& Combined approach & \textbf{76.80} & \textbf{75.67} & \textbf{77.84} & 79.87& -\\
\cline{2-8}
{}&\multirow{4}{*}{VPU}& Source only& \textbf{23.67} & - & - & -& -\\
{}&{}& Fine-tuning only & - & \textbf{51.51} & 46.50 & 67.47& \textbf{69.62}\\
{}&{}& Cross-domain pseudo-label only & 02.67 & 02.86 & 02.68 & 02.77& -\\
{}&{}& Combined approach & 02.66 & 02.68 & 02.67 & 02.66& -\\
\hline
\multirow{12}{*}{Pratheepan}&\multirow{4}{*}{SFA}& Source only & 80.66 & - & - & -& -\\
{}&{}& Fine-tuning only & - & \textbf{93.68} & \textbf{94.70} & \bf{95.69}& 95.99\\
{}&{}& Cross-domain pseudo-label only & 82.50 & 83.36 & 83.63 & 90.60& -\\
{}&{}& Combined approach & \textbf{82.96} & 84.12 & 84.47 & 92.93& -\\
\cline{2-8}
{}&\multirow{4}{*}{Compaq}& Source only & \textbf{63.49} & - & - & -& -\\
{}&{}& Fine-tuning only & - & 64.88 & 66.10 & 68.97& \textbf{70.52}\\
{}&{}& Cross-domain pseudo-label only & 39.50 & 41.26 & 44.69 & 62.39& -\\
{}&{}& Combined approach & 34.72 & 36.22 & 39.05 & 57.06& -\\
\cline{2-8}
{}&\multirow{4}{*}{VPU}& Source only & \textbf{36.68} & - & - & -& -\\
{}&{}& Fine-tuning only & - & \textbf{51.61} & \textbf{60.19} & \textbf{68.15}& \textbf{69.44}\\
{}&{}& Cross-domain pseudo-label only & 02.66 & 02.66 & 02.67 & 02.77& -\\
{}&{}& Combined approach & 02.65 & 02.66 & 02.67 & 02.74& -\\
\hline
\end{tabular}
\label{tab:6}
\end{table*}

% SFA 60.76% melhor, Pratheepan 14.94% melhor, mas VPU 34.96% PIOR
\begin{figure}[tb]
  \centering
    \includegraphics[width=.48\textwidth]{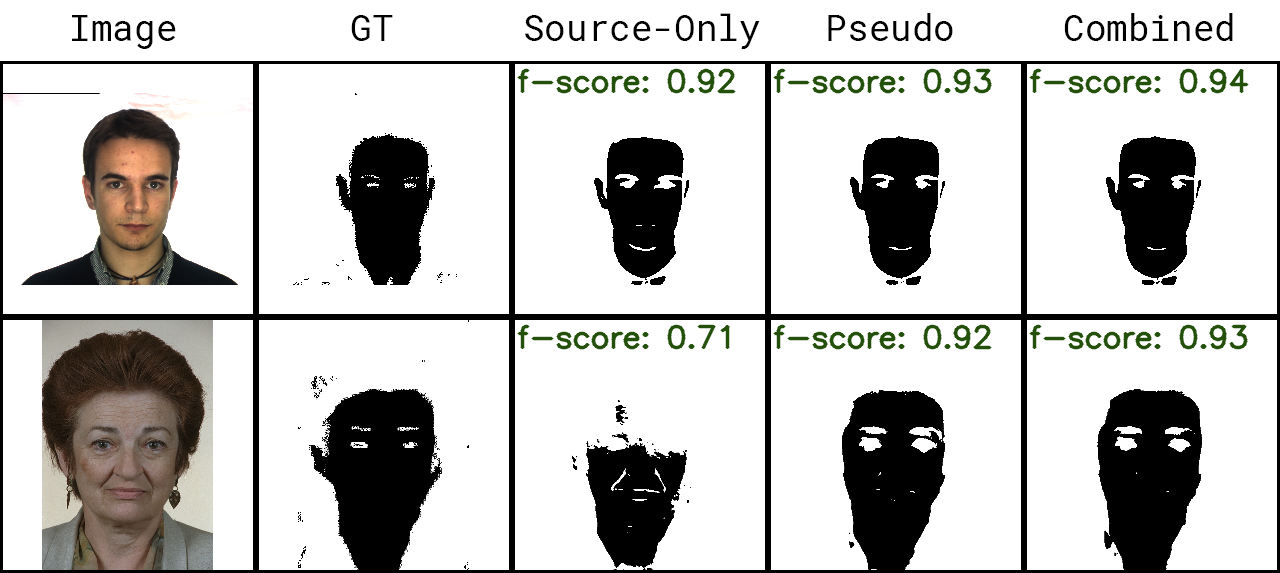}
  \caption{Domain adaptation from Compaq to SFA using no real labels from target. From left to right: target test image, ground truth and results with source only, domain adaptation based on cross-domain pseudo-labels and the combined domain adaptation + transfer learning approach.}
  \label{fig:compaq-sfa}
\end{figure}

\begin{figure}[tb]
  \centering
    \includegraphics[width=.48\textwidth]{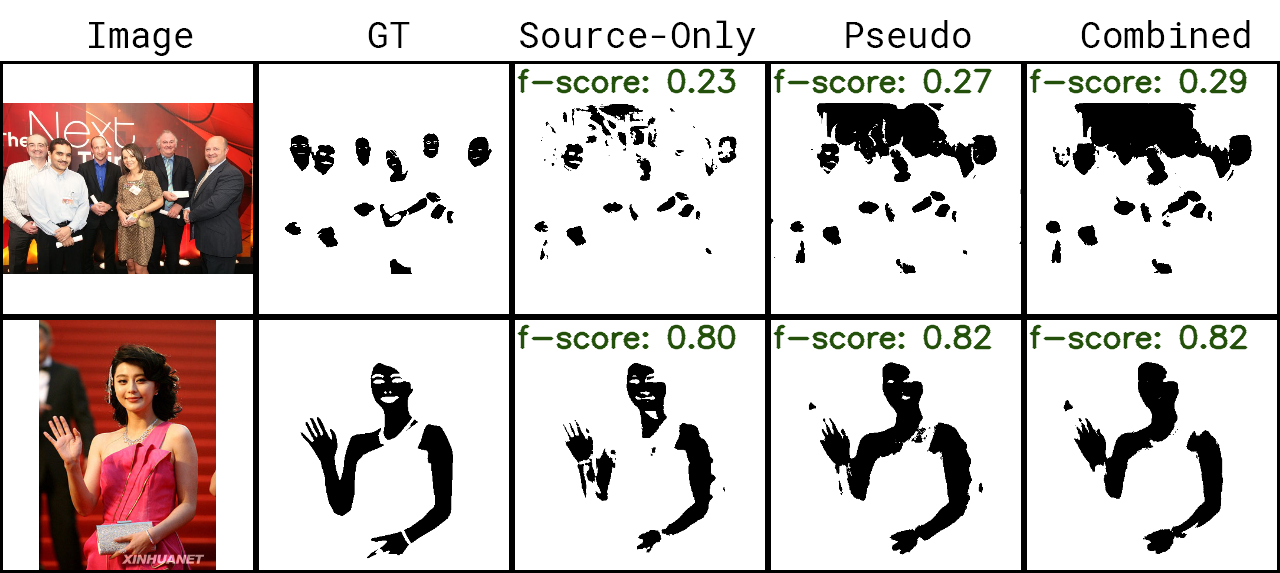}
  \caption{Domain adaptation from Compaq to Pratheepan using no real labels from target (same setting as Figure~\ref{fig:compaq-sfa}).}
  \label{fig:compaq-pratheepan}
\end{figure}

Compaq has confirmed our expectations of being the most generalizable source dataset, not only for being the most numerous in terms of sample images but also due to their diversity in appearance. 
The use of Compaq as source lead to very good results on SFA and Pratheepan as targets. 
These results are illustrated in figures \ref{fig:compaq-sfa} and \ref{fig:compaq-pratheepan}, respectively, which show the effects of using different domain adaptation methods with no labels from target dataset. Note that when using Compaq as source and Pratheepan as target, the gain of the domain adaptation approaches is very expressive when compared to target only training. domain adaptation methods got better results using any amount of labels on the target training set, being the combined approach the best option in most cases.  Using 50\% of training data our cross-domain pseudo-label approach was better than regular supervised training with 100\% of training data. Besides that, all the results of domain adaptation methods with no labels were better than the state-of-the-art results of color-based approaches presented in Section~\ref{sec:indomain}.

\begin{figure}[t]
  \centering
    \includegraphics[width=.48\textwidth]{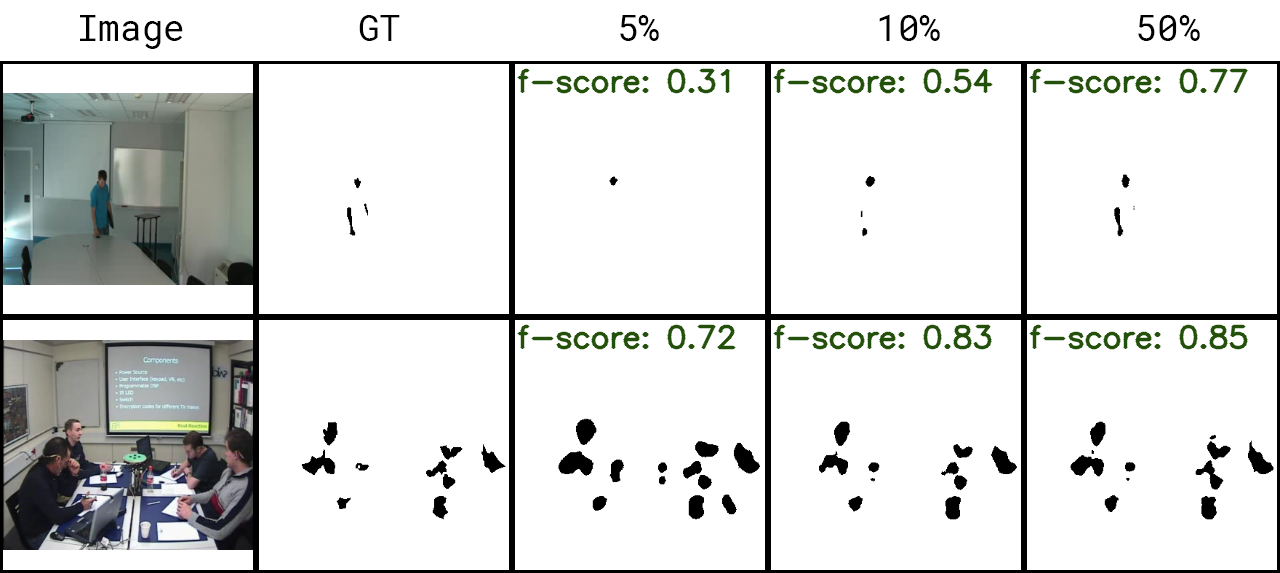}
  \caption{Adaptation from Pratheepan to VPU with fine-tuning TL. From left to right: target test image, ground truth and resutls with 5, 10 and 50\% of labels on the target training set.}
  \label{fig:pratheepan-vpu}
\end{figure}

When VPU is the target dataset, Pratheepan outperformed Compaq as source dataset. However, the pseudo-labels caused negative transfer, leading to very bad results when domain adaptation was used. The results with fine-tuning were better than regular supervised training with all evaluated amounts of training labels. In this scenario, the reference color-based approach by \cite{SANMIGUEL20132102} was beaten starting from 10\% of training label usage. Results with 5, 10 and 50\% are shown for two sample images in Figure~\ref{fig:pratheepan-vpu}.

\begin{figure}[t]
  \centering
    \includegraphics[width=.48\textwidth]{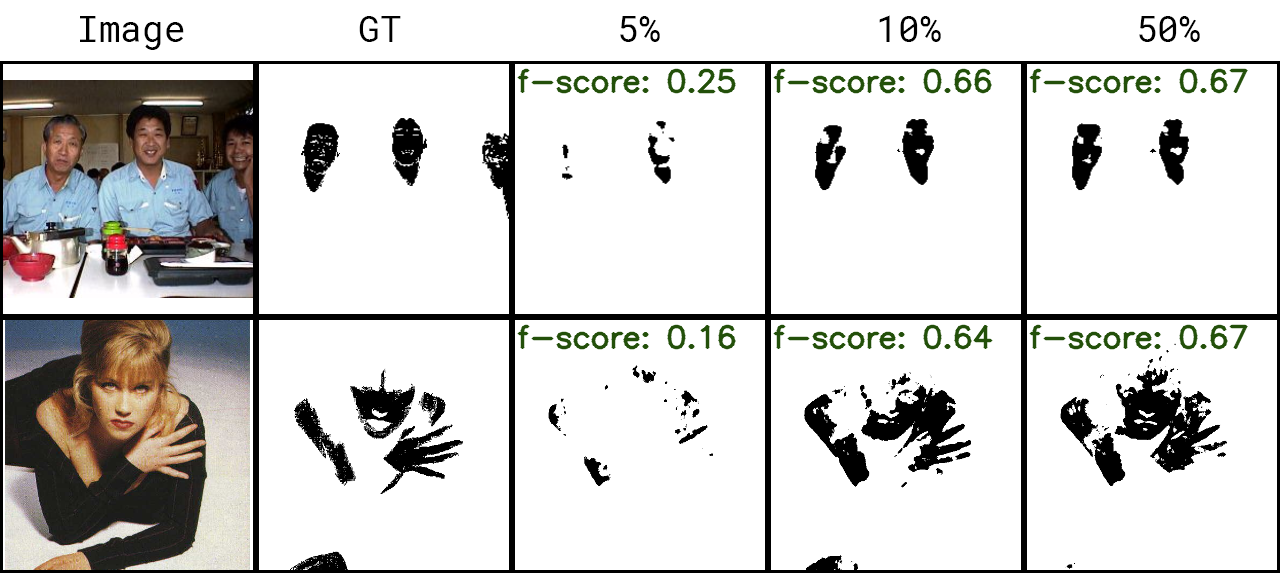}
  \caption{Adaptation from Pratheepan to Compaq with fine-tuning using different amounts of labels on the target training set (following the same setting as Figure~\ref{fig:pratheepan-vpu}).}
  \label{fig:pratheepan-compaq}
\end{figure}

Still with Pratheepan as source dataset, but with Compaq as target, the ``source only'' result was reasonable and surpassed the color-based approach. However, we observed that domain adaptation methods did not remarkably improve the results from regular supervised training. Figure \ref{fig:pratheepan-compaq} shows the results of fine-tuning from Pratheepan to Compaq.

\subsection{Discussion}
Although most approaches for skin detection in the past have assumed that skin regions are nearly textureless \cite{face_chai1999,huynh-thu_skin-color-based_2002,Phumg2002,shaik_comparative_2015,BRANCATI201733}, our results give the unintuitive conclusion that texture and context play an important role.
A holistic segmentation approach like fully convolutional networks, taking the whole image as input, in conjunction with adequate domain adaptation methods, has more generalization power than local approaches like color and patch-based. The improvement level and best domain adaptation approach varies depending on how close target and source domains are and on the diversity of the samples in the source dataset. The closer the domains and the higher the source variety, the higher the improvement.
%
% Our proposed cross-domain pseudo-label and combined approaches have showed to be very effective when the source domain has more diversity and somehow includes images similar to the target domain (compaq $\rightarrow$ sfa, for example).
For example, a very positive transfer from Compaq$\rightarrow$SFA was observed because Compaq is more diverse and includes samples whose appearance is somewhat similar to those of SFA.
This is intuitive, as these approaches depend on the quality of the pseudo-labels.
When the transition between domains goes from specific to diverse datasets, the pseudo-labels are expected to be of low quality, thus, not contributing to the target model training. On these situations, fine-tuning has showed to be more effective, although with the drawback of requiring at least some few labeled images for training.

\begin{figure}[tb]
  \centering
    \includegraphics[width=.48\textwidth]{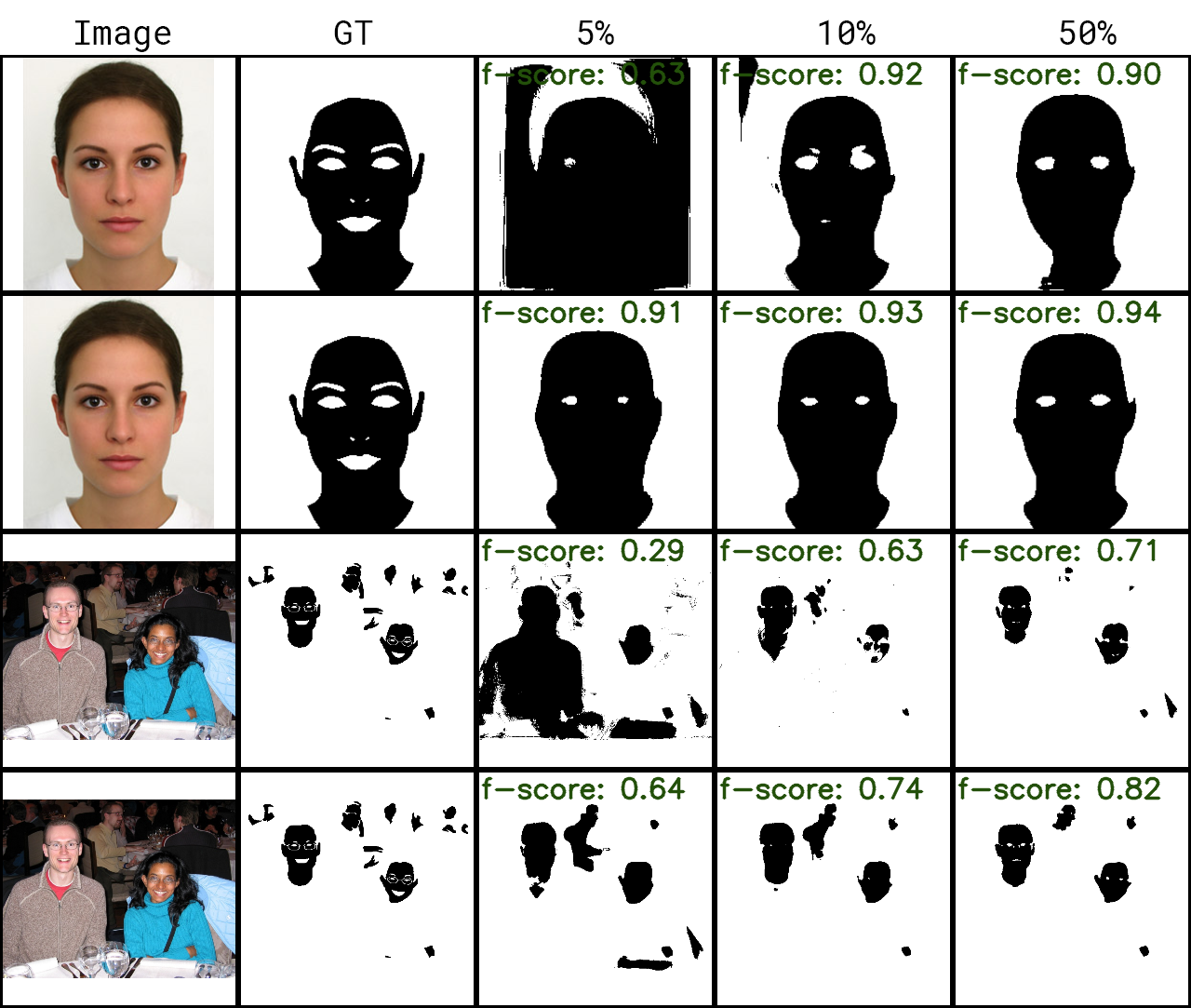}
  \caption{Comparison of source only vs.\ domain adaptation combined approach in the Compaq$\rightarrow$Pratheepan scenario with different proportions of labeled target training samples. For each target test image, the first row is regular supervised training and the second is the combined domain adaptation approach.}
  \label{fig:so-pseudo}
\end{figure}

Domain Adaptation methods have also showed improvements when compared to regular supervised training in cases where the target has few images, like Pratheepan and VPU.
The level of improvement depends on the amount of labeled target training data and on the similarity of source and target domains.
The higher the amount, the lower the improvement, and the higher the similarity, the higher the improvement.
Figure \ref{fig:so-pseudo} shows a comparison of regular supervised training versus the combined approach in the Compaq$\rightarrow$Pratheepan scenario with 5, 10 and 50\% of the target training samples with labels. This scenario is good for the pseudo-label approach, since Compaq has more diversity than Pratheepan.
Note the superiority of combined approach in all levels of target 
labels availability. 

\begin{figure}[htb]
  \centering
    \includegraphics[width=.48\textwidth]{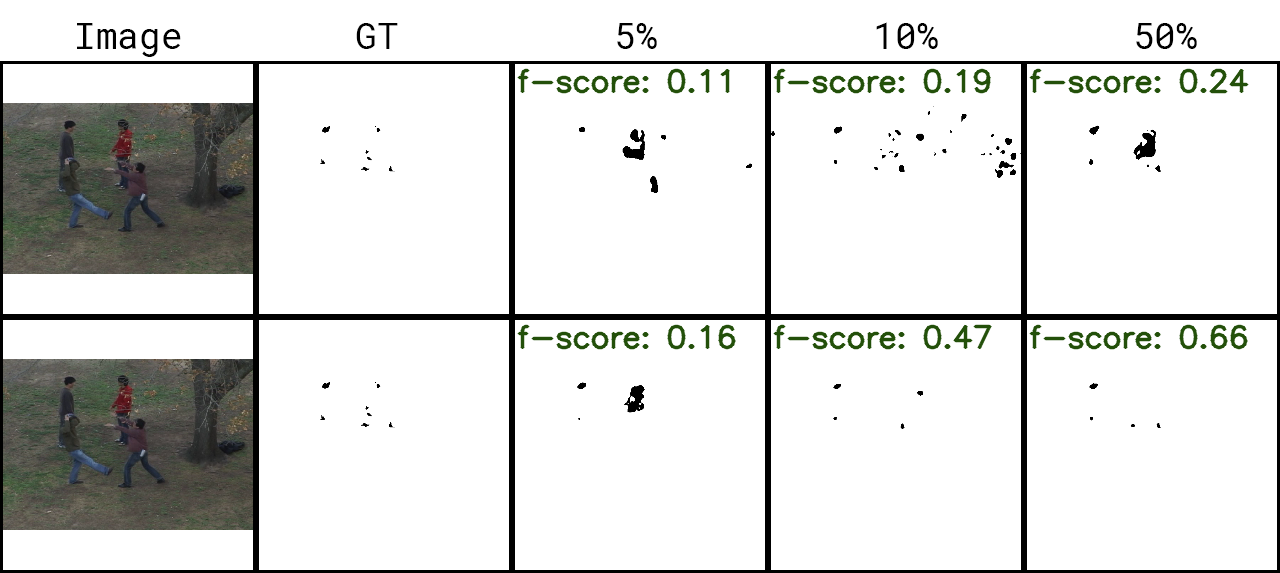}
  \caption{Comparison of source only vs.\ fine-tune in the Pratheepan $\rightarrow$ VPU scenario with different proportions of labeled target training samples. For each target test image, the first row is regular supervised training and the second is the fine-tuning approach.}
  \label{fig:so-fine}
\end{figure}

Figure \ref{fig:so-fine}, on the other hand, shows the  comparison of regular supervised training versus the fine-tune approach in the Pratheepan $\rightarrow$ VPU scenario. As Pratheepan does not cover scenes that occur on VPU, the fine-tune approach perform better than cross-domain pseudo-labels in this scenario.

Another important aspect to be addressed is the criticism for the applicability of CNN approaches to real-time applications. 
The criticism is probably valid for patch-based CNN approaches, but it does not hold for our FCN holistic approach. The average prediction time of our patch-based CNN, using a simple Nvidia GTX-1080Ti, with a frame size of $768\times768$ pixels, is 7 seconds per image which is indeed not suitable for real-time applications. However, our U-Net prediction time is 80 ms per frame for the same setup, i.e., 12.5 images are processed per second (without parallel processing).
\cite{BRANCATI201733} has reported prediction time of about 10ms per frame with frame size of $300\times400$ pixels ($8\times$ faster on images that are $5\times$ smaller), which is indeed a bit faster, at a penalty of producing worse results. 

\section{Conclusions}
In this work we refuted some common criticisms regarding the use of Deep Convolutional Networks for skin segmentation. We compared two CNN approaches (patch-based and holistic) to the state-of-the-art pixel-based solutions for skin detection in in-domain situations. As our main contribution, we proposed novel approaches for semi-supervised and unsupervised domain adaptation applied to skin segmentation using CNNs and evaluated it with a extensive set of experiments.

Our evaluation of in-domain skin detection approaches on different domains/datasets showed the expected and incontestable superiority of CNN based approaches over color based ones.  Our U-Net model obtained F$_1$ scores which were on average 30\% %29.99\%
better than the state-of-the-art recent published color based results. In more homogeneous and clean datasets, like SFA, our F$_1$ score was 73\% %73.43\% 
better. Even in more difficult and heterogeneous datasets, like Prathepaan and VPU, our U-Net CNN was more than 10\% better.

More importantly, we experimentally came to the unintuitive conclusion that a holistic approach like U-net, besides being much faster, gives better results than a patch-based local approach.

We also concluded that the common critique of lack of generalization of CNNs does not hold true against our experimental data. With no labeled data on the target domain, our domain adaptation method's F$_1$ score is an improvement of 60\% over color-based results %59.97\%
for homogeneous target datasets like SFA and 13\% %13.11\%
in heterogeneous datasets like Pratheepan.  

Note that the approaches for both inductive transfer learning (TL) and unsupervised domain adaptation (DA) are baseline methods. More sophisticated approaches have been proposed for both problems,
such as \cite{JDA_ICCV2013,farajidavar_etal_ATTM_bookchapter2017,csurka_comprehensive_2017,Gatys_etal_StyleTransfer_cvpr2016}.
Our study shows that, despite the simplicity of the chosen methods, they greatly contribute to the improvement in the performance on skin segmentation across different datasets, showing that even better results are expected with more sophisticated methods. For example, our results were in general better than the individual methods gathered in \cite{Lumini2018FairCO} and on par with their proposed ensemble method. %, being 5\% and 9\% worse on Pratheepan and Compaq, respectively, and 7\% better on SFA.

Among the possible directions for future work we propose the use of iterative domain adaptation methods which progressively improve pseudo-labels \cite{farajiDavar_etal_vectar2011,JDA_ICCV2013,farajidavar_etal_ATTM_bookchapter2017}. Another possibility is to exploit a metric that compares the distribution of source and target samples in order to avoid negative transfer by automatically suggesting whether to use DA, TL, the combined approach or if it is better to disregard the source domain and use only the target samples (when labeled samples are available). A similar idea has been used by \cite{FarajiDavar-BMVC-2014}, but with a different goal: automatic selection of classifiers for transfer learning.
The use of GANs-based methods for domain adaptation \cite{Tzeng2017} is also a promising avenue for future work.

\subsection{Code and weights availability}
All source code developed to perform the training and the evaluations, along side the resulting models weights will be made available from \url{http://cic.unb.br/~teodecampos/} upon acceptance of the paper.

\section*{Acknowledgments}
Dr.\ de Campos acknowledges the support of CNPq fellowship PQ 314154/2018-3.